\documentclass{article} 
\usepackage{iclr2027_conference,times}

\usepackage{amsmath,amsfonts,bm}

\def\eqref#1{equation~\ref{#1}}

\def\1{\bm{1}}

\DeclareMathAlphabet{\mathsfit}{\encodingdefault}{\sfdefault}{m}{sl}
\SetMathAlphabet{\mathsfit}{bold}{\encodingdefault}{\sfdefault}{bx}{n}

\usepackage{hyperref}
\usepackage{url}
\usepackage{amssymb}

\usepackage{amsmath}
\usepackage{amsthm}

\usepackage{booktabs}
\usepackage{multirow}
\usepackage{graphicx}
\usepackage{wrapfig}

\usepackage{subcaption}
\usepackage{colortbl}
\usepackage[table]{xcolor}

\usepackage{fancyhdr}

\title{ADT: All-Atom Diffusion Tokenizer for 3D Molecular Structures}

\author{
Xuantuo Huang\textsuperscript{1}, Encheng Su\textsuperscript{2}, Aoran Wang\textsuperscript{3}, Pan Tan\textsuperscript{3}, Shengyong Xu\textsuperscript{1}, Yizhou Wang\textsuperscript{4}, 
\\
\textbf{
Chen Tang\textsuperscript{4}, Shixiang Tang\textsuperscript{4$\dagger$}, Mingxiao Li\textsuperscript{3$\dagger$}
}
\\[4pt]
\textsuperscript{1}Peking University
\textsuperscript{2}University of Science and Technology of China
\textsuperscript{3}Shanghai AI Laboratory
\\
\textsuperscript{4}The Chinese University of Hong Kong
\\
\textsuperscript{$\dagger$}Corresponding authors
\vspace{-0.9cm}
}

\iclrfinalcopy 
\begin{document}

\maketitle
\pagestyle{fancy}

\fancyhf{}

\begin{abstract}
Discrete structural tokenization provides a promising route to autoregressive 3D molecular generation.
Yet, existing tokenizers face a trade-off between compact representation and all-atom reconstruction: atom-level tokens preserve atomic details but produce long sequences, while compact residue-level protein tokens typically omit side-chain geometry.
To overcome this limitation, we propose \textbf{A}ll-atom \textbf{D}iffusion \textbf{T}okenizer (\textbf{ADT}), which decouples token granularity from reconstruction resolution.
ADT uses one structural token per atom for small molecules and one token per protein residue that jointly represents backbone and side-chain geometry.
To mitigate geometric distortion introduced by quantization, we leverage the generative capacity of diffusion models to recover fine-grained atomic geometry beyond deterministic decoding from discrete tokens.
Extensive experiments across small molecules and proteins demonstrate that ADT achieves strong reconstruction from compact discrete representations while enabling high-quality autoregressive generation of both small molecules and all-atom protein structures.
These results establish compact structural tokens as an effective discrete interface for all-atom molecular reconstruction and generation.
\end{abstract}

\section{Introduction}

Generating three-dimensional molecular structures is a central task in drug discovery and protein engineering~\citep{su2026lamgen,lin2025silico,luo2026towards,huang2024learning}. Molecular interactions depend critically on atomic geometry, from small-molecule conformations to protein backbones and side chains~\citep{jin2018junction,irwin2022chemformer}. 
Many existing approaches directly model molecular structures in continuous coordinate space~\citep{hoogeboom2022equivariant,xu2023geometric,dunn2024mixed,morehead2024geometry}.
While effective, continuous coordinate representations do not naturally fit the autoregressive framework of large language models.
Structural tokenization instead maps 3D geometry into discrete representations, enabling molecular structures to be generated autoregressively as token sequences and decoded into atomic coordinates.
This formulation allows molecular structure generation to leverage scalable language-modeling paradigms demonstrated across language and biomolecular modeling~\citep{kaplan2020scaling,hoffmann2022training,bhatnagar2026scaling}.

\begin{wrapfigure}{r}{0.62\textwidth}
    \centering
    \includegraphics[width=\linewidth]{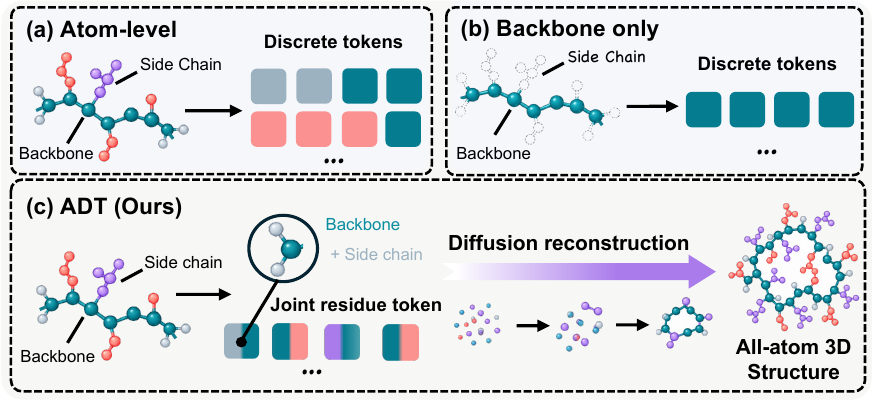}
        \setlength{\abovecaptionskip}{-7pt}
    \caption{
Illustration of tokenizer granularity.
\textit{(a)} Atom-level tokenization  assigns one token to each atom.
\textit{(b)} Backbone-only tokenization omits side-chain geometry.
\textit{(c)} ADT jointly encodes backbone and side-chain geometry into one token per residue and reconstructs the all-atom structure.
}
\vspace{-10pt}
    \label{fig:teaser}
\end{wrapfigure}

Recent molecular tokenizers encode geometry at different granularities, as shown in Fig.~\ref{fig:teaser}.
Atom-level tokenizers~\citep{liu2025biotoken,geikyan2026accurate} assign structural codes to individual atoms, preserving all-atom coverage with sequence lengths that scale with atom count.
For proteins, this yields substantially longer sequences than residue-level representations and requires more autoregressive prediction steps.
Backbone-based protein tokenizers~\citep{wang2025dplm,hayes2025simulating,gao2025foldtoken} use compact residue-level codes but do not explicitly encode side-chain conformations, limiting faithful reconstruction of the original all-atom structure.
Recent approaches further incorporate generative decoders to improve backbone reconstruction~\citep{dilip2026flow} and enable adaptive compression
of C$\alpha$ structures~\citep{dilip2026adaptive}.  

These limitations raise two challenges for compact all-atom tokenization. 
First, \emph{compact all-atom encoding}: side-chain conformations are essential for describing molecular interactions, yet encoding every atom separately produces long token sequences for proteins.
A useful representation must retain these atomic details while controlling sequence length.
Second, \emph{accurate reconstruction under quantization error}: mapping continuous conformations to discrete codes loses geometric information, making it difficult to recover precise local geometry and overall molecular structure.

We address these challenges by introducing \textbf{A}ll-atom \textbf{D}iffusion \textbf{T}okenizer (ADT), which
decouples token granularity from reconstruction resolution.
For proteins, we jointly encode backbone and side-chain geometry into one structural token per residue.
To mitigate geometric distortion introduced by quantization, we formulate reconstruction as conditional generative modeling, exploiting the generative capacity of a diffusion decoder to recover fine-grained atomic geometry from discrete structural tokens and chemical identities.

Specifically, ADT encodes all-atom geometry into one structural token per chemical unit: an atom for small molecules and a residue for proteins. 
To make these tokens better suited for generative modeling, we introduce a bitwise mutual-information regularizer that promotes effective code utilization.
Conditioned on these tokens and their chemical identities, a diffusion decoder reconstructs atomic geometry from the discrete bottleneck.
For proteins, the decoder first models residue-level context and then lifts each residue representation to atom-level features, enabling all-atom reconstruction from a single token per residue.
Beyond reconstruction, we use the learned tokens for autoregressive molecular generation, where the chemical identity sequence is generated first, followed by the corresponding structural tokens conditioned on the complete chemical sequence.
Together, these designs enable compact molecular tokens to support all-atom reconstruction and generation without tying token granularity to atomic resolution.

We comprehensively evaluate our framework on reconstruction and generation across small molecules and proteins.
Our experiments show that ADT achieves accurate reconstruction from compact discrete representations, including all-atom protein reconstruction from a single structural token per residue.
Beyond reconstruction, our experiments demonstrate that the learned tokens enable high-quality autoregressive generation of both small molecules and all-atom protein structures.
Our main contributions are summarized as follows:
1)  We introduce a unified discrete representation for small molecules and proteins that decouples token granularity from reconstruction resolution, using one structural token per atom for small molecules and per residue for proteins.
2)  We formulate reconstruction as conditional generation and introduce a coarse-to-fine diffusion decoder that recovers fine-grained atomic geometry from discrete structural tokens.
3)  Our experiments demonstrate that compact structural tokens support reconstruction and autoregressive generation of both small molecules and all-atom protein structures.

\section{Related Work}

\subsection{Molecular Tokenization}
Molecular geometry can be serialized through coordinate strings or invariant geometric sequences~\citep{flam2023language,zholus2025bindgpt,li2025geometry}, while learned tokenizers compress continuous geometry into discrete structural codes. Such representations have been explored for small molecules~\citep{gao2026tokenizing,geikyan2026accurate} and protein structures, including VQ-based backbone tokenization~\citep{gaujac2024learning}, ESM3~\citep{hayes2025simulating}, DPLM-2~\citep{wang2025dplm}, and ProToken~\citep{lin2025unifying}. Recent methods further extend structural coverage beyond protein backbones: Bio2Token uses atom-level quantized representations across biomolecular systems~\citep{liu2025biotoken}, while BiHiTo introduces hierarchical structural codes~\citep{zheng2026bihito}. At residue granularity, SLAE learns compact local all-atom representations~\citep{chen2026slae}, and related backmapping methods reconstruct atomistic structures from compressed representations~\citep{han2026constraint}.
We instead formulate molecular tokenization over chemical units, pairing each atom or residue identity with a discrete structural code and decoupling token granularity from all-atom reconstruction resolution, thereby providing a common discrete interface across molecular systems.

\subsection{3D Molecular Generation}

3D molecular generation has been explored in both continuous coordinate space and discrete token space. Early autoregressive approaches construct molecular structures sequentially by predicting atom types and coordinates~\citep{gebauer2019symmetry,gebauer2022inverse}, while diffusion and flow models directly generate molecular geometry for small molecules~\citep{hoogeboom2022equivariant,peng2023moldiff,dunn2024mixed} and protein structures~\citep{YimTBMDBJ23,watson2023novo,chu2024all,geffner2026proteina,li2026proteinae}.
More recently, discrete structural representations bridge 3D geometry and language modeling by converting molecular geometry into token sequences for autoregressive generation. This paradigm has been explored for small molecules~\citep{gao2026tokenizing} and proteins~\citep{gaujac2024learning,hayes2025simulating,wang2025dplm,dilip2026flow,dilip2026adaptive}. However, discrete protein generation largely operates at C$\alpha$ or backbone resolution, omitting side-chain and local atomic details needed to represent interaction interfaces and functional sites~\citep{chu2024all}. This limits its applicability to structure-based design tasks that require atomic-level detail.
We address this limitation by combining compact autoregressive structural tokens with generative all-atom decoding.
\section{Preliminaries}

\paragraph{3D Molecular Representation.}
A 3D molecular structure is represented as a sequence of $L$ structural units,
$
\mathcal{M}=(\mathcal{U}_1,\ldots,\mathcal{U}_L),
$
where each structural unit $\mathcal{U}_i$ contains a set of atoms indexed by
$\mathcal{A}_i$, with Cartesian coordinates
$\mathbf{x}_{i,j}\in\mathbb{R}^{3}$ for $j\in\mathcal{A}_i$.
For small molecules, each structural unit corresponds to a single atom,
whereas for proteins, each structural unit corresponds to a residue containing
multiple atoms.
We denote the chemical identities of the structural units by
$C=(c_1,\ldots,c_L)$, corresponding to atom types for small molecules
and amino-acid types for proteins.
The atomic geometry is represented as
$\mathbf{X}\in\mathbb{R}^{L\times A\times3}$, where $A$ denotes the number of
atomic identities per structural unit.

\paragraph{Discrete Molecular Tokenization.}
Given a molecular structure $\mathcal{M}$, a discrete molecular tokenizer maps
it into continuous latent representations through an encoder $f_{\theta}$,
followed by a quantization function $Q$ that converts the latent representations
into a sequence of discrete tokens:
\begin{equation}
\mathbf{H}=f_{\theta}(\mathcal{M}), \qquad
\mathbf{S}=Q(\mathbf{H})=(s_1,\ldots,s_L),
\quad s_i\in\{0,\ldots,K-1\},
\end{equation}
where $K$ denotes the size of the discrete vocabulary.
A decoder $g_{\phi}$ reconstructs the atomic geometry from the discrete structural tokens $\mathbf{S}$,
conditioned on the structural-unit identities $\mathbf{C}$.

\begin{figure}[t]
    \centering
    \includegraphics[width=0.95\linewidth]{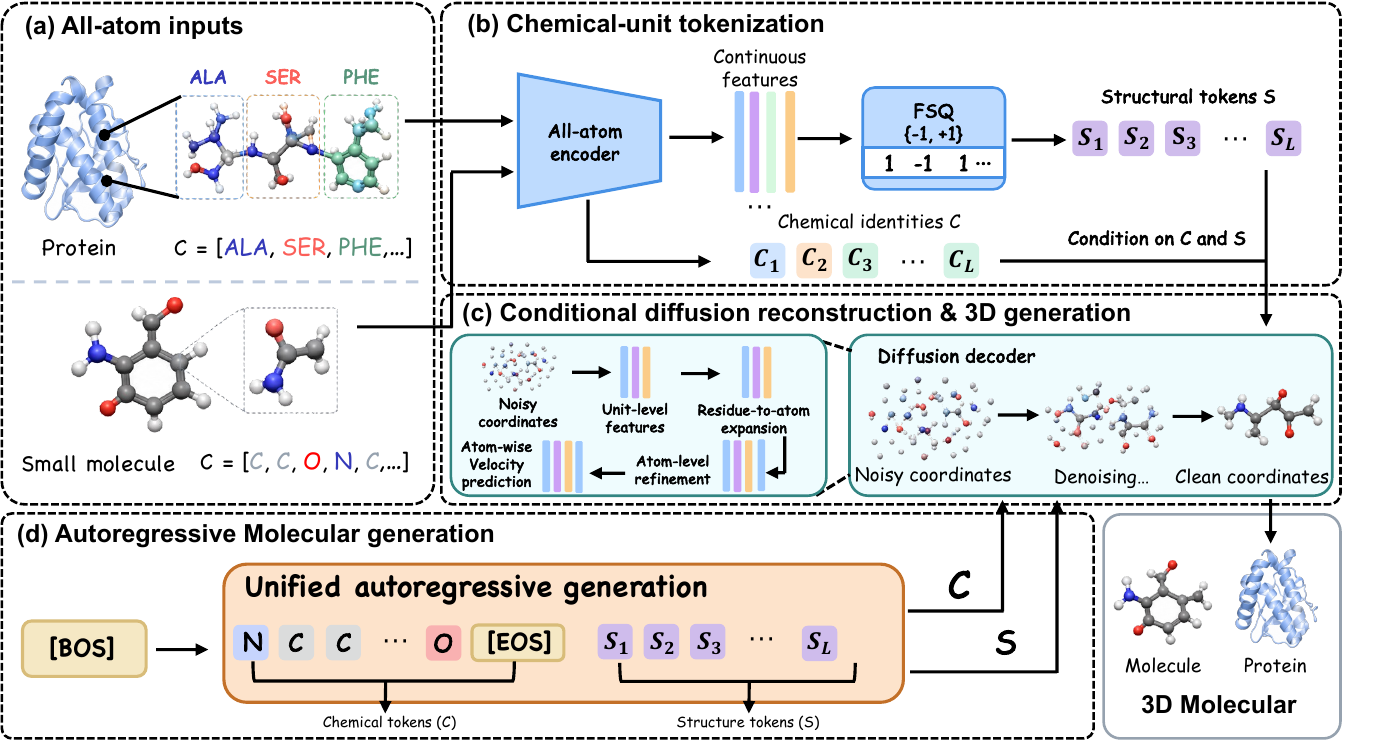}
\caption{
Overview of ADT.
\textit{(a)} All-atom structures are represented at chemical-unit granularity.
\textit{(b)} Each unit is mapped to a discrete structural token via FSQ.
\textit{(c)} Conditional diffusion reconstructs all-atom geometry, with residue-to-atom expansion for proteins.
\textit{(d)} ADT enables autoregressive generation of all-atom molecular structures.
}
    \label{fig:techincal}
\end{figure}

\section{Methodology}
As shown in Fig.~\ref{fig:techincal}, we introduce All-atom Diffusion Tokenizer (ADT), which decouples token granularity from reconstruction resolution. 
Each small-molecule atom and protein residue is represented by a single structural token, while a diffusion decoder mitigates quantization-induced distortion to recover fine-grained all-atom geometry.
This design provides a common discrete interface across molecular modalities while retaining all-atom reconstruction.

\subsection{Unified Molecular Structure Tokenization}

\paragraph{Unified Structure Representation.}
Small molecules and proteins exhibit different structural granularities, with atom-level units for small molecules and residue-level units for proteins. Following the structural-unit representation defined above, we jointly embed the chemical identity and atomic geometry of each structural unit into a common sequence space:
\begin{equation}
\mathbf{H}^{0}
=
\operatorname{MLP}
\left(
[\mathbf{X};\mathbf{M}]
\right)
+
\mathbf{E}(\mathbf{C}),
\qquad
\mathbf{H}
=
\operatorname{TransformerEnc}(\mathbf{H}^{0})
\in\mathbb{R}^{L\times d}.
\end{equation}
Here, $\mathbf{M}$ is a binary mask indicating valid atoms, and $\mathbf{E}(\mathbf{C})$ denotes the structural-unit identity embeddings, corresponding to atom types for small molecules and amino-acid types for proteins. This formulation yields one contextual representation per structural unit while preserving its underlying atomic geometry.

\paragraph{Discrete Molecular Tokenization.}
We further map the unified continuous representations into a common discrete
space using finite scalar quantization (FSQ)~\citep{mentzer2024finite}.
Specifically, given the contextual representation $\mathbf{H}$, we project
each structural unit into an $m$-dimensional quantization space, independently
quantize each dimension into two symmetric levels, and map the resulting binary
code to a discrete token:
\begin{equation}
    \mathbf{Z}
    =
    \operatorname{MLP}(\mathbf{H})
    \in\mathbb{R}^{L\times m},
    \qquad
    \widehat{z}_{i,j}
    =
    Q(z_{i,j})\in\{-1,+1\},
    \qquad
    s_i
    =
    \sum_{j=1}^{m}
    \frac{\widehat{z}_{i,j}+1}{2}\,2^{j-1}.
\end{equation}

This yields a structural token sequence
$\mathbf{S}=(s_1,\ldots,s_L)$ over the same vocabulary of size $K=2^m$,
providing a common discrete representation for structural units across
molecular modalities.

\subsection{Molecular Structure Reconstruction}

\paragraph{Coarse-to-Fine Diffusion Reconstruction.}
We reconstruct all-atom geometry from discrete structural tokens through coarse-to-fine diffusion decoding. The decoder progressively lifts token-level representations to atom-level representations, decoupling token granularity from reconstruction resolution.
Given the discrete structural tokens $\mathbf{S}$, we recover their quantized
representations $\widehat{\mathbf{Z}}$ and combine them with the corresponding
chemical identity embeddings to form the conditioning representations:
\begin{equation}
    \mathbf{D}
    =
    \operatorname{MLP}(\widehat{\mathbf{Z}})
    +
    \mathbf{E}_{\mathrm{dec}}(\mathbf{C})
    \in\mathbb{R}^{L\times d},
    \qquad
    \mathbf{P}_t
    =
    \mathcal{D}(\mathbf{X}_t;\mathbf{D},t)
    \in\mathbb{R}^{L\times d},
\end{equation}
where $\mathbf{E}_{\mathrm{dec}}$ is a learnable decoder-side embedding of the chemical identities $\mathbf{C}$, and $\mathbf{D}$ denotes the resulting conditioning representations. $\mathbf{X}_t$ denotes the noisy molecular coordinates at diffusion time $t$, which are projected into structural-unit representations and processed by the diffusion Transformer $\mathcal{D}$. The resulting $\mathbf{P}_t$ contains one contextual representation for each structural unit.

Since each protein token represents an entire residue, directly decoding all atoms from a single residue-level representation creates a bottleneck for fine-grained atomic reconstruction.
We therefore adopt a coarse-to-fine strategy that expands each intermediate residue representation into $A$ atom-wise representations, conditioned on learnable atom embeddings and noisy coordinates:
\begin{equation}
\begin{gathered}
    \mathbf{U}_{t,i,a}^{0}
    =
    \mathbf{W}\mathbf{P}_{t,i}
    +
    \mathbf{e}_{a}
    +
    \operatorname{MLP}_{1}(\mathbf{x}_{t,i,a}),
    \\
    \mathbf{U}_{t,i}
    =
    \operatorname{Transformer}(\mathbf{U}_{t,i}^{0}),
    \qquad
    \widehat{\mathbf{v}}_{t,i,a}
    =
    \operatorname{MLP}_{2}(\mathbf{U}_{t,i,a})
    \in\mathbb{R}^{3}.
\end{gathered}
\end{equation}
Here, $\mathbf{e}_a$ is a learnable embedding that distinguishes individual atoms within each residue. 
The remaining decoder layers refine the expanded atom-wise representations, which are finally projected to atom-wise denoising velocities.
This transition from $\mathbb{R}^{L\times d}$ to $\mathbb{R}^{L\times A\times d}$ decouples residue-level tokenization from all-atom reconstruction resolution.

\paragraph{Reconstruction Objective.}
We train each diffusion decoder using a velocity-prediction objective.
Conditioned on the discrete structural representation, the diffusion decoder predicts the velocity of each atom from its noisy coordinates. For a molecule with $N$ atoms, the objective is:
\begin{equation}
    \mathcal{L}_{\mathrm{recon}}
    =
    \mathbb{E}_{t,\boldsymbol{\epsilon}}
    \left[
        \frac{1}{N}
        \sum_{i=1}^{N}
        \left\|
            \widehat{\mathbf{v}}_{t,i}
            -
            \mathbf{v}_{t,i}
        \right\|_2^2
    \right],
\end{equation}
where $\widehat{\mathbf{v}}_{t,i}$ and $\mathbf{v}_{t,i}$ denote the predicted
and target velocities of the $i$-th atom, respectively. This objective
propagates atom-level reconstruction supervision through the discrete
bottleneck, encouraging each token to retain the structural information
required for recovering molecular geometry.

\paragraph{Mutual-Information Regularization.}
High reconstruction fidelity alone does not guarantee effective generative modeling in the discrete space. We therefore introduce a bitwise mutual-information regularizer to encourage diverse code utilization across structural units while maintaining confident individual assignments. 
For the $j$-th pre-quantization scalar $z_{i,j}$ of structural unit $\mathcal{U}_i$, we define the soft binary assignment:
\begin{equation}
p_{i,j}
=
q(B_j=1\mid \mathbf{h}_i)
=
\frac{\exp(z_{i,j}/\tau)}
{\exp(z_{i,j}/\tau)+\exp(-z_{i,j}/\tau)},
\end{equation}
where $B_j$ denotes the $j$-th binary quantization variable and $\tau$ is the temperature.
Using the binary Shannon entropy $h(p)=-p\log p-(1-p)\log(1-p)$,  we estimate the normalized bitwise mutual information over all valid structural units in each mini-batch as:
\begin{equation}
\mathcal{L}_{\mathrm{MI}}
=
- \frac{1}{m\log 2}
\sum_{j=1}^{m}
\left[
h\left(\mathbb{E}_{i}[p_{i,j}]\right)
-
\mathbb{E}_{i}\left[h(p_{i,j})\right]
\right].
\end{equation}
The first term encourages diverse bit utilization across structural units, while the second encourages confident assignments for individual units. The resulting $\mathcal{L}_{\mathrm{MI}}$ is used as a regularization term during tokenizer training.

\paragraph{Protein Geometry Regularization.}
For proteins, we complement the reconstruction objective with a backbone loss
$\mathcal{L}_{\mathrm{bb}}$ and auxiliary geometric losses on bond lengths,
bond angles, steric clashes, and $\mathrm{C}_{\alpha}$ pairwise distances.
The resulting training objective is:
\begin{equation}
\mathcal{L}_{\mathrm{prot}}
=
\mathcal{L}_{\mathrm{recon}}
+
\lambda_{\mathrm{MI}}\mathcal{L}_{\mathrm{MI}}
+
\boldsymbol{\omega}^{\top}
[
\mathcal{L}_{\mathrm{bb}},
\mathcal{L}_{\mathrm{bond}},
\mathcal{L}_{\mathrm{angle}},
\mathcal{L}_{\mathrm{clash}},
\mathcal{L}_{\mathrm{dist}}
]^{\top}.
\end{equation}
where
$\boldsymbol{\omega}\in\mathbb{R}^{5}$ balances the five
auxiliary terms.  For small-molecule tokenizer, the training objective is $\mathcal{L}_{\mathrm{recon}}
+ \lambda_\mathrm{MI} \mathcal{L}_{\mathrm{MI}}.$
Detailed definitions are provided in Appendix~\ref{app:training}.

\subsection{Autoregressive Molecule Generation }
\label{sec:dual_ar}

Discrete structural tokens allow both chemical identity and 3D geometry to be generated within the same autoregressive sequence-modeling paradigm.  
Building on this representation, we factorize molecular generation into chemical sequence generation followed by structural token generation conditioned on the complete chemical sequence.
Let $\mathbf{C}=(c_1,\ldots,c_L,\mathrm{EOS})$ and $\mathbf{S}=(s_1,\ldots,s_L)$ denote the chemical and structural sequences, respectively. We factorize their joint distribution as:
\begin{equation}
\begin{aligned}
p_\theta(\mathbf{C},\mathbf{S})
=p_\theta(\mathbf{C})~p_\theta(\mathbf{S}\mid\mathbf{C})
=
\left[\prod_{i=1}^{L+1}p_\theta(c_i\mid\mathbf{C}_{<i})\right]
\left[\prod_{i=1}^{L}p_\theta(s_i\mid\mathbf{C},\mathbf{S}_{<i})\right].
\end{aligned}
\end{equation}
Thus, each  structural prediction has access to chemical information from all molecular positions. Both autoregressive processes are trained jointly with teacher forcing:
\begin{equation}
\mathcal{L}_{\mathrm{AR}}
=
-\sum_{i=1}^{L+1}\log p_\theta(c_i\mid\mathbf{C}_{<i})
-\beta\sum_{i=1}^{L}\log p_\theta(s_i\mid\mathbf{C},\mathbf{S}_{<i}).
\end{equation}
where $\beta$ balances the chemical and structural autoregressive objectives.

At inference, chemical tokens are sampled until $\mathrm{EOS}$, determining the sequence length $L$. The model then generates one structural token for each chemical position, and the molecular decoder converts $(\mathbf{S},\mathbf{C})$ into 3D coordinates $\mathbf{X}$. For structure generation from a given chemical sequence, $\mathbf{C}$ is fixed and only the structural autoregression is performed.
\section{Experiments}

We evaluate ADT on both small molecules and proteins, covering 3D structure reconstruction and autoregressive generation.
We report reconstruction and generation results, followed by ablation studies on the key design choices of ADT.

\subsection{Experiment Setup}
\paragraph{Datasets.}
We conduct experiments on benchmarks spanning small molecules and proteins. For small molecules, we use QM9~\citep{ramakrishnan2014quantum}, containing approximately 130K organic molecules with up to nine heavy atoms, and $\nabla^2$DFT~\citep{khrabrov2024nabla}, using 1.9M molecules for training and the three test splits for evaluation on unseen molecules with 8–27 heavy atoms. For proteins, we use AFDB SwissProt~\citep{varadi2022alphafold}, with CASP14~\citep{kryshtafovych2021critical}/CASP15~\citep{kryshtafovych2023critical} held out for protein reconstruction evaluation. We exclude all evaluation targets from the tokenizer training set
and remove training proteins with at least 80\% sequence identity
to evaluation target, as measured over Foldseek~\citep{van2024fast}.
Further details are provided in Appendix~\ref{app:implementation}.

\paragraph{Evaluation Metrics.} Small-molecule reconstruction is evaluated using structure match, validity, and RMSD. For protein reconstruction, we report C$\alpha$-lDDT, backbone and all-atom RMSD, all-atom lDDT, TM-score, GDT-TS, and GDT-HA. Small-molecule generation is evaluated using validity, uniqueness, diversity, and PoseBusters validity, while protein generation metrics are specified in the corresponding experiments.
Detailed definitions and evaluation protocols are provided in Appendix~\ref{app:metirc}.

\paragraph{Training and Evaluation.}
We use a codebook size of 8,192 and Transformer-based encoders with diffusion decoders for both molecular modalities. The tokenizers are optimized with AdamW with learning rate $3 \times 10^{-4}$ on a single NVIDIA H200 GPU. Detailed model architectures, optimization settings, loss weights, and evaluation protocols are provided in Appendix~\ref{app:model}.

\subsection{Reconstruction on small molecules and all-atom proteins}

\textbf{Reconstruction on small molecules.}
We evaluate molecular reconstruction  on QM9 and the three
\begin{wraptable}{r}{0.48\textwidth}
\vspace{-10pt}
\centering
\caption{Reconstruction results on QM9.}
\label{tab:qm9_reconstruction}
\resizebox{\linewidth}{!}{%
\begin{tabular}{llcc}
\toprule
\textbf{Type} & \textbf{Method} 
& \textbf{Match} $\uparrow$ 
& \textbf{RMSD} $\downarrow$ \\
\midrule

\multirow{3}{*}{Continuous}
& GeoLDM~\citep{xu2023geometric} & 87.2 & 0.1820 \\
& ADiT~\citep{joshi2025allatom} & 94.6 & 0.0890 \\
& SinAE~\citep{ren2026sinae} & 100.0 & 0.0007 \\

\midrule

\multirow{3}{*}{Discrete}
& Mol-StrucTok~\citep{gao2026tokenizing} & 90.7 & 0.1280 \\
& CoordToken~{\hypersetup{linkbordercolor={0 1 0}}%
\hyperlink{cite.geikyan2026accurate}{Geikyan et al. (2026)}}
& 99.4 & 0.0478 \\
& \textbf{ADT (Ours)} & \textbf{99.7} & \textbf{0.0343} \\
\bottomrule
\end{tabular}%
}
\vspace{-10pt}
\end{wraptable}
$\nabla^2$DFT test splits, covering both in-distribution molecules and generalization to unseen structures and scaffolds. 
Tables~\ref{tab:qm9_reconstruction} and~\ref{tab:nabladft_reconstruction} reveal three main results:
1) Our method consistently outperforms existing discrete tokenizers. Compared with CoordToken, it reduces 
reconstruction RMSD by 28.2\% on QM9 and 23.8–25.6\% across the three $\nabla^2$DFT test splits, while achieving a higher match rate on QM9 and maintaining 100\% validity on $\nabla^2$DFT.
2) Discrete tokenization does not preclude high reconstruction accuracy. On QM9, our method surpasses the continuous GeoLDM and ADiT baselines in both match rate and RMSD.
3) The advantage extends beyond QM9 to larger and more diverse molecules. Across the $\nabla^2DFT$ test splits, including unseen structures and out-of-distribution scaffolds, ADT reduces RMSD over existing discrete tokenizers by 23.8–25.6\% while maintaining complete validity.

\begin{table*}[t]
\centering
\caption{Reconstruction performance on the $\nabla^2$DFT benchmark.}
\label{tab:nabladft_reconstruction}

\resizebox{0.92\textwidth}{!}{%
\begin{tabular}{l|cc|cc|cc}
\toprule
\multirow{2}{*}{\textbf{Method}}
& \multicolumn{2}{c|}{\textbf{Conformer}}
& \multicolumn{2}{c|}{\textbf{Structure}}
& \multicolumn{2}{c}{\textbf{Scaffolds}} \\
& \textbf{RMSD} $\downarrow$ & \textbf{Valid} $\uparrow$
& \textbf{RMSD} $\downarrow$ & \textbf{Valid} $\uparrow$
& \textbf{RMSD} $\downarrow$ & \textbf{Valid} $\uparrow$ \\
\midrule

mol2token~\citep{liu2025biotoken}
& 0.2010 & 47.1
& 0.2010 & 47.1
& 0.2010 & 47.1 \\

CoordToken~{\hypersetup{linkbordercolor={0 1 0}}%
\hyperlink{cite.geikyan2026accurate}{Geikyan et al. (2026)}}
& 0.0671 & 100.0
& 0.0673 & 100.0
& 0.0671 & 100.0 \\

\textbf{ADT (Ours)}
& \textbf{0.0499} & \textbf{100.0}
& \textbf{0.0512} & \textbf{100.0}
& \textbf{0.0511} & \textbf{100.0} \\

\bottomrule
\end{tabular}%
}
\end{table*}

\paragraph{All-atom protein reconstruction.}
\label{exp:all-atom protein reconstruction}
Table~\ref{tab:protein_reconstruction} shows that residue-level tokenization can enable accurate all-atom reconstruction while preserving backbone fidelity.
Since DPLM-2 reconstructs only protein backbones, we recover all atoms from the reconstructed backbone using PDBFixer, cg2all~\citep{heo2024one}, and Rosetta~\citep{leaver2011rosetta3}. We additionally evaluate cg2all with the input backbone fixed, denoted as cg2all (fixed).
Compared with DPLM-2 followed by all-atom completion, our method substantially reduces all-atom and backbone RMSD on CASP14 and CASP15, while maintaining strong backbone lDDT. Notably, each residue is represented by only a single structural token yet decoded into its full atomic geometry. These results demonstrate that compact token granularity need not constrain reconstruction resolution, supporting our design of decoupling residue-level encoding from all-atom decoding.

\begin{table*}[t]
\centering
\caption{Protein reconstruction performance on CASP14 and CASP15.
DPLM-2 variants recover all-atom structures using different backbone-to-all-atom reconstruction methods.}
\label{tab:protein_reconstruction}

\resizebox{\textwidth}{!}{%
\begin{tabular}{clccccccc}
\toprule
&
\textbf{Method}
& \textbf{C$\alpha$-lDDT} $\uparrow$
& \textbf{BB RMSD} $\downarrow$
& \textbf{AA lDDT} $\uparrow$
& \textbf{AA RMSD} $\downarrow$
& \textbf{TM-score} $\uparrow$
& \textbf{GDT-TS} $\uparrow$
& \textbf{GDT-HA} $\uparrow$ \\
\midrule

\multirow{6}{*}{\rotatebox[origin=c]{90}{\textbf{CASP14}}}
& DPLM-2~\citep{wang2025dplm}
& 0.8820 & 2.8576 & -- & -- & 0.8860 & 0.7718 & 0.5724 \\

& \quad + Rosetta fixbb
& 0.8820 & 2.8576 & 0.7532 & 3.4918 & 0.8859 & 0.7717 & 0.5723 \\

& \quad + PDBFixer
& 0.8820 & 2.8576 & 0.6632 & 3.9657 & 0.8859 & 0.7717 & 0.5723 \\

& \quad + cg2all (nofix)
& 0.8820 & 2.8093 & 0.7710 & 3.4120 & 0.8859 & 0.7718 & 0.5722 \\

& \quad + cg2all (fixed)
& 0.8819 & 2.8576 & 0.7611 & 3.4358 & 0.8859 & 0.7717 & 0.5723 \\

& \textbf{ADT (Ours)}
& \textbf{0.8970}
& \textbf{0.9628}
& \textbf{0.8223}
& \textbf{1.3157}
& \textbf{0.9695}
& \textbf{0.9398}
& \textbf{0.7686} \\

\midrule

\multirow{6}{*}{\rotatebox[origin=c]{90}{\textbf{CASP15}}}
& DPLM-2~\citep{wang2025dplm}
& 0.8933 & 4.0223 & -- & -- & 0.8644 & 0.7032 & 0.5104 \\

& \quad + Rosetta fixbb
& 0.8933 & 4.0223 & 0.7671 & 4.5931 & 0.8644 & 0.7032 & 0.5104 \\

& \quad + PDBFixer
& 0.8933 & 4.0223 & 0.6767 & 5.0213 & 0.8644 & 0.7032 & 0.5104 \\

& \quad + cg2all (nofix)
& 0.8933 & 3.9796 & 0.7879 & 4.5037 & 0.8644 & 0.7029 & 0.5101 \\

& \quad + cg2all (fixed)
& 0.8933 & 4.0224 & 0.7772 & 4.5288 & 0.8644 & 0.7032 & 0.5104 \\

& \textbf{ADT (Ours)}
& \textbf{0.8972}
& \textbf{1.5191}
& \textbf{0.8258}
& \textbf{1.8261}
& \textbf{0.9470}
& \textbf{0.9001}
& \textbf{0.7285} \\

\bottomrule
\end{tabular}%
}
\end{table*}

\subsection{Generation on small molecules and all-atom proteins}

\textbf{Generation on small molecules.}
We evaluate unconditional molecular generation on QM9.
As shown in Table~\ref{tab:qm9_generation}, ADT outperforms the evaluated discrete baselines across all four metrics, improving Valid and PB Valid over Mol-StrucTok from 0.97/0.88 to 1.00/0.94.
ADT is also competitive with continuous generative models, matching the best PB Valid while maintaining high uniqueness and diversity.
Notably, SinAE achieves substantially lower reconstruction error but comparable generation performance, indicating that reconstruction fidelity alone does not determine generative quality.
Overall, these results demonstrate that ADT provides a strong discrete representation for autoregressive 3D molecular generation.

\begin{table}[h]
\centering
\caption{Unconditional molecular generation results on QM9.}
\label{tab:qm9_generation}
\small
\begin{tabular}{llcccc}
\toprule
\textbf{Type} & \textbf{Method} 
& \textbf{Valid} $\uparrow$ 
& \textbf{Unique} $\uparrow$ 
& \textbf{Div.} $\uparrow$ 
& \textbf{PB Valid} $\uparrow$ \\
\midrule
\multirow{5}{*}{Continuous}
& EQGAT-diff~\citep{le2024navigating} & 0.99 & 1.00 & 0.89 & 0.94 \\
&SemlaFlow~\citep{irwin2024efficient} &0.99 &0.95 &0.89&0.94\\
& FlowMol~\citep{dunn2024mixed}    & 0.97 & 0.97 & 0.92 & 0.94 \\
& ADiT~\citep{joshi2025allatom}       & 0.97 & 0.97 & 0.90 & 0.92 \\
& SinAE~\citep{ren2026sinae}      & 0.97 & 0.97 & 0.91 & 0.94 \\
\midrule
\multirow{3}{*}{Discrete}
& Mol-StrucTok~\citep{gao2026tokenizing}  & 0.97 & 0.85 & 0.87 & 0.88 \\
& CoordToken~{\hypersetup{linkbordercolor={0 1 0}}%
\hyperlink{cite.geikyan2026accurate}{Geikyan et al. (2026)}}   & 0.95 & 0.93 & 0.90 & 0.86 \\
& \textbf{ADT (Ours)}     & \textbf{1.00} & \textbf{0.96} & \textbf{0.91} &\textbf{0.94} \\
\bottomrule
\end{tabular}
\vspace{-4pt}
\end{table}

\paragraph{Sequence-Conditioned Protein Generation.}
We train Qwen3.5-0.8B~\citep{team2026qwen3} on tokenized AFDB structures for
sequence-conditioned protein structure generation, with training details provided
in Appendix~\ref{app:sequence-details}.
To isolate the effect of structural tokenization, we use the same Qwen3.5-0.8B
backbone with either the DPLM-2 tokenizer or ADT, while keeping all other
components fixed.
Given an amino-acid sequence, the model autoregressively predicts one structural
token per residue.
ADT directly decodes these tokens into all-atom coordinates using the frozen
diffusion decoder, whereas DPLM-2 outputs are further converted to all-atom
structures using the side-chain reconstruction methods described in
Section~\ref{exp:all-atom protein reconstruction}.

We evaluate on 183 CAMEO22 targets using sequence-length matching,
C$\alpha$-lDDT, backbone lDDT, TM-score, and all-atom lDDT.
As shown in Table~\ref{tab:protein_generation_postprocess}, ADT achieves an all-atom lDDT
of 0.543, outperforming all post-processed DPLM-2 variants, including cg2all (nofix)
at 0.503.
ADT also improves C$\alpha$-lDDT from 0.5607 to 0.5740 and TM-score from
0.633 to 0.665, while maintaining exact sequence-length matching.
These results show that a single residue-level structural token can jointly
represent backbone and side-chain geometry, enabling direct all-atom generation
without a separate side-chain reconstruction stage.

\begin{table}[t]
\centering
\caption{Sequence-conditioned protein structure generation on CAMEO22.}
\label{tab:protein_generation_postprocess}
\small
\setlength{\tabcolsep}{6pt}
\begin{tabular}{lccccc}
\toprule
\textbf{Method}
& \textbf{Length Match}
& \textbf{C$\alpha$-lDDT}
& \textbf{BB-lDDT}
& \textbf{TM-score}
& \textbf{AA-lDDT} \\
\midrule
DPLM-2          & 1.0 & 0.5607 & 0.5683 & 0.633 & -- \\
+ Rosetta fixbb & 1.0 & 0.5607 & 0.5683 & 0.633 & 0.482 \\
+ PDBFixer      & 1.0 & 0.5607 & 0.5683 & 0.633 & 0.372 \\
+ cg2all nofix  & 1.0 & 0.5607 & 0.5683 & 0.633 & 0.503 \\
+ cg2all fixed  & 1.0 & 0.5607 & 0.5683 & 0.633 & 0.496 \\
\midrule
\textbf{ADT (Ours)}
& \textbf{1.0}
& \textbf{0.5740}
& \textbf{0.5801}
& \textbf{0.665}
& \textbf{0.543} \\
\bottomrule
\end{tabular}
\vspace{-12pt}
\end{table}

\subsection{Effect of All-Atom Joint Encoding}

We ablate how backbone and side-chain geometry should be represented by comparing three encoding strategies under the same residue-level tokenization framework: encoding backbone geometry only, encoding backbone and side chains separately (Separate BB+SC), and jointly encoding all-atom geometry into a single structural token per residue. 
Separate BB+SC independently encodes backbone and side-chain geometry into two residue-level structural tokens, whereas our joint variant represents both with a single token.
All variants use the same model architecture and training protocol, differing only in the structural information encoded by the discrete representation. Table~\ref{tab:protein_encoding_ablation} reveals three key observations: 
1) \textbf{Geometric Complementarity}: Joint all-atom encoding consistently outperforms backbone-only encoding across all metrics, while achieving comparable side-chain reconstruction to separate encoding and stronger overall backbone fidelity, revealing complementary geometric information between backbone and side chains.
2) \textbf{Improved Structural Fidelity}: Compared with separate encoding, joint encoding better preserves global structural fidelity, as reflected by consistently higher C$\alpha$-lDDT, TM-score, GDT-TS, and GDT-HA.
3) \textbf{Compact Representation}: A single residue-level discrete token enables high-fidelity all-atom reconstruction, decoupling token granularity from reconstruction resolution.

\begin{table*}[t]
\centering
\caption{Ablation of protein structural encoding strategies on CASP14 and CASP15.}
\label{tab:protein_encoding_ablation}

\resizebox{\textwidth}{!}{%
\renewcommand{\arraystretch}{1.35}
\begin{tabular}{clcccccccc}
\toprule
& \textbf{Method}
& \textbf{C$\alpha$-lDDT} $\uparrow$
& \textbf{BB RMSD} $\downarrow$
& \textbf{AA lDDT} $\uparrow$
& \textbf{AA RMSD} $\downarrow$
& \textbf{SC RMSD} $\downarrow$
& \textbf{TM-score} $\uparrow$
& \textbf{GDT-TS} $\uparrow$
& \textbf{GDT-HA} $\uparrow$ \\
\midrule

\multirow{3}{*}{\rotatebox[origin=c]{90}{\textbf{CASP14}}}
& Backbone only
& 0.8802
& 1.2804
& 0.7699
& 1.9269
& 2.3858
& 0.9555
& 0.9038
& 0.7241 \\

& Separate BB+SC
& 0.8762
& 1.3105
& \textbf{0.8262}
& 1.5075
& 1.6763
& 0.9549
& 0.9106
& 0.7238 \\

& \textbf{Joint All-atom (ADT)}
& \textbf{0.8970}
& \textbf{0.9628}
& 0.8223
& \textbf{1.3157}
& \textbf{1.5938}
& \textbf{0.9695}
& \textbf{0.9398}
& \textbf{0.7686} \\

\midrule

\multirow{3}{*}{\rotatebox[origin=c]{90}{\textbf{CASP15}}}
& Backbone only
& 0.8864
& 1.6517
& 0.7866
& 2.2386
& 2.6672
& 0.9365
& 0.8875
& 0.7150 \\

& Separate BB+SC
& 0.8827
& \textbf{1.4884}
& \textbf{0.8306}
& \textbf{1.7711}
& \textbf{2.0341}
& 0.9416
& 0.8896
& 0.7065 \\

& \textbf{Joint All-atom (ADT)}
& \textbf{0.8972}
& 1.5191
& 0.8258
& 1.8261
& 2.0779
& \textbf{0.9470}
& \textbf{0.9001}
& \textbf{0.7285} \\

\bottomrule
\end{tabular}%
}
\vspace{-12pt}
\end{table*}

\subsection{Analysis and Ablation Studies}

\begin{figure*}[t]
    \centering

    \begin{minipage}[t]{0.40\textwidth}
        \centering
        \includegraphics[width=\linewidth]{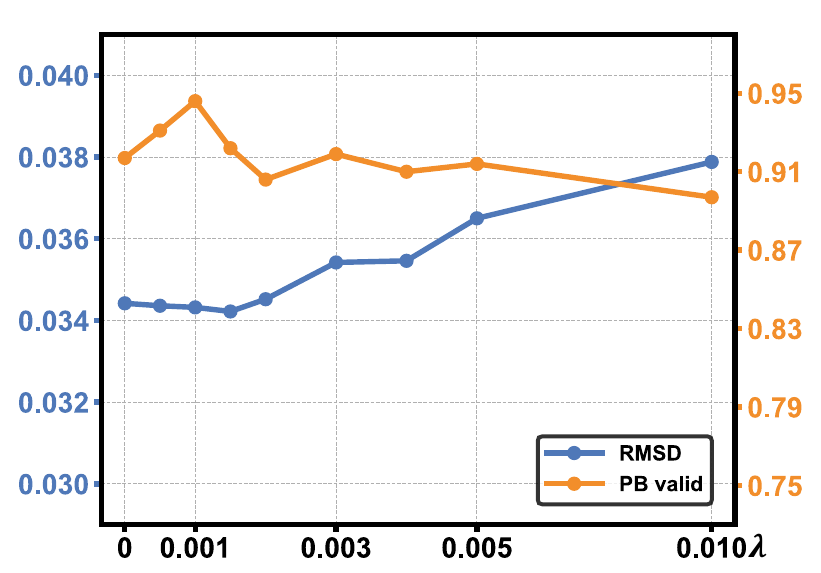}
        \caption{Effect of MI regularization on reconstruction and generation.}
        \label{fig:infomax}
    \end{minipage}
    \hfill
    \begin{minipage}[t]{0.58\textwidth}
        \centering
        \includegraphics[width=\linewidth]{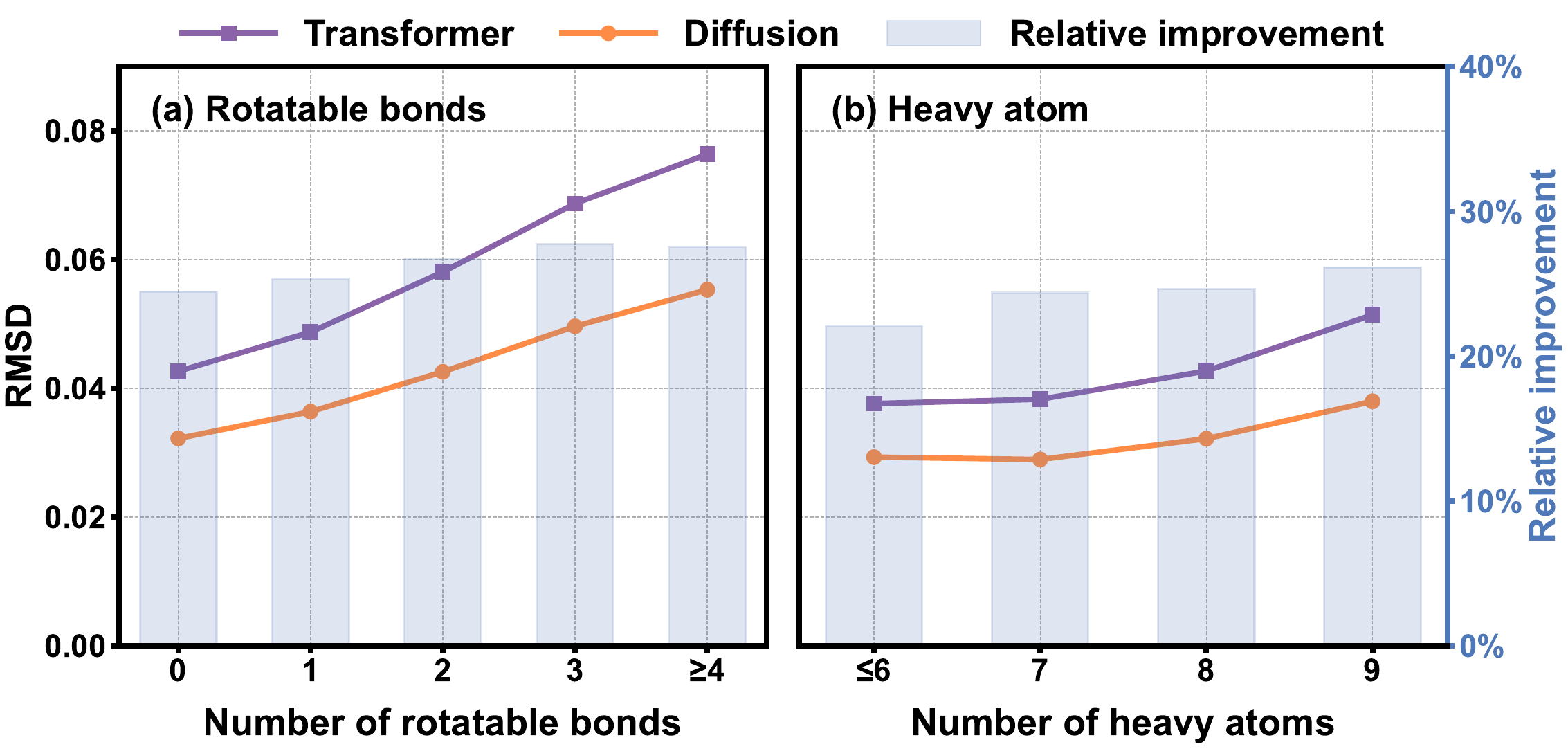}
        \captionof{figure}{
            Reconstruction gains from generative decoding across molecular complexity.
        }
        \label{fig:decoder_complexity}
    \end{minipage}
\vspace{-10pt}
\end{figure*}

\paragraph{Effect of Bitwise Regularization.}
We investigate the effect of bitwise regularization on reconstruction
and downstream generation.
As shown in Fig.~\ref{fig:infomax}, mild regularization improves PB Valid,
suggesting that shaping the discrete representation with the bitwise
mutual-information objective facilitates downstream autoregressive modeling.
Both PB Valid and reconstruction RMSD remain relatively stable across a
broad range of $\lambda$, indicating limited sensitivity to the exact
regularization strength.
At larger $\lambda$, reconstruction RMSD gradually increases without
further improvements in PB Valid, motivating the use of moderate
regularization.

\paragraph{Ablating Diffusion Reconstruction.}
To isolate the contribution of generative decoding, we replace the conditional diffusion decoder with a Transformer that directly predicts atomic coordinates, while keeping the remaining architectural and training settings unchanged.
As shown in Table~\ref{tab:decoder_depth}, diffusion decoding consistently outperforms the direct Transformer decoder across different depths, with the improvement increasing as the number of decoder layers grows. This trend further supports the effectiveness of diffusion decoding in mitigating quantization-induced geometric distortion.

We further stratify the QM9 test set by the number of heavy atoms and rotatable bonds to examine how this benefit scales with molecular complexity, as illustrated in Fig.~\ref{fig:decoder_complexity}. Diffusion decoding yields larger absolute reductions in reconstruction error as complexity increases, while maintaining broadly comparable relative improvements. The growing absolute advantage on more challenging molecules suggests that generative decoding becomes increasingly valuable as the geometric information discarded by discrete quantization becomes harder to recover.

\begin{table}[h]
\centering
\caption{Effect of decoder depth on QM9 reconstruction.}
\label{tab:decoder_depth}
\small
\begin{tabular}{lccccc}
\toprule
\textbf{Decoder} & \textbf{Layers}
& \textbf{Match} $\uparrow$
& \textbf{RMSD $\downarrow$}
& \textbf{PB Valid $\uparrow$}
& \textbf{RMSD Gain $\uparrow$} \\
\midrule
Transformer & 4  & 97.5 & 0.0695 & 0.88 & \multirow{2}{*}{\textbf{27.19\%}} \\
Diffusion   & 4  & \textbf{98.3} & \textbf{0.0506} & \textbf{0.89} & \\
\midrule
Transformer & 8  & 98.4 & 0.0636 & 0.90 & \multirow{2}{*}{\textbf{33.18\%}} \\
Diffusion   & 8  & \textbf{99.1} & \textbf{0.0425} & \textbf{0.91} & \\
\midrule
Transformer & 12 & 98.9 & 0.0584 & 0.92 & \multirow{2}{*}{\textbf{41.27\%}} \\
Diffusion   & 12 & \textbf{99.7} & \textbf{0.0343} & \textbf{0.93} & \\
\bottomrule
\end{tabular}
\end{table}

\section{Conclusion}

We introduced All-atom Diffusion Tokenizer (ADT), a discrete tokenizer for 3D molecular structures that decouples token granularity from reconstruction resolution. ADT represents small molecules with one structural token per atom and proteins with one token per residue, while retaining all-atom reconstruction through conditional diffusion decoding. By leveraging generative decoding to mitigate geometric information loss introduced by quantization, ADT achieves accurate reconstruction from compact discrete representations. The resulting tokens further support autoregressive generation of both small molecules and all-atom protein structures. Together, our results demonstrate that compact structural tokens can serve as an effective discrete interface for all-atom molecular modeling, bridging efficient token-based generation with fine-grained atomic geometry.

\newpage
\bibliographystyle{iclr2027_conference}
\bibliography{iclr2027_conference}

\clearpage
\appendix

\begin{center}
    {\LARGE\bfseries ADT: All-Atom Diffusion Tokenizer for 3D Molecular Structures\par}

    \vspace{1.2cm}

    \noindent\rule{0.14\linewidth}{0.6pt}
    \hspace{0.2em}
    {\LARGE\bfseries Appendix}
    \hspace{0.2em}
    \rule{0.14\linewidth}{0.6pt}
\end{center}

\vspace{0.8cm}

\noindent The appendix is organized as follows:

\vspace{0.4cm}

\begin{itemize}
    \setlength{\itemsep}{1.2em}

    \item Appendix~\ref{app:further_experiments} provides further experimental results, including fine-grained reconstruction analyses, comparisons across token granularities and discrete protein tokenizers, and qualitative results.

    \item Appendix~\ref{app:additional_ablations} presents additional ablation studies on structural information in discrete tokens, coarse-to-fine decoding, discrete capacity, and autoregressive factorization.

    \item Appendix~\ref{app:implementation} outlines additional implementation details, including model architectures, training settings, and evaluation metrics.

    \item Appendix~\ref{app:limitations}  contains the limitations of our work.

    \item Appendix~\ref{app:llm} describes the use of large language models during manuscript preparation.

\end{itemize}

\section{Further Experimental Results}
\label{app:further_experiments}
\subsection{Comparison Across Token Granularities}

Atom-level tokenization provides direct coverage of all-atom protein geometry, but requires one structural token per atom, resulting in substantially longer sequences than residue-level tokenization. On CASP14 and CASP15, proteins contain on average 2,141.3 and 2,598.5 atoms but only 268.6 and 336.5 residues, respectively, requiring approximately \textbf{8.0$\times$} and \textbf{7.7$\times$} as many structural tokens as  residue-level tokenization. Table~\ref{tab:protein_allatom_reconstruction} compares all-atom reconstruction across these token granularities. Atom-level tokenizers achieve lower reconstruction error, whereas ADT directly reconstructs complete atomic geometry from a single structural token per residue and substantially improves over residue-level backbone tokenization followed by external all-atom completion.

\begin{table}[h]
\centering
\caption{All-atom protein reconstruction across token granularities on CASP14 and CASP15.}
\label{tab:protein_allatom_reconstruction}

\resizebox{\columnwidth}{!}{%
\begin{tabular}{clcccc}
\toprule
& \multirow{2}{*}{\textbf{Model}}
& \multicolumn{2}{c}{\textbf{CASP14}}
& \multicolumn{2}{c}{\textbf{CASP15}} \\
\cmidrule(lr){3-4}
\cmidrule(lr){5-6}
&
& \textbf{AA RMSD $\downarrow$}
& \textbf{TM-score $\uparrow$}
& \textbf{AA RMSD $\downarrow$}
& \textbf{TM-score $\uparrow$} \\
\midrule

\multirow{2}{*}{\textbf{Atom-level}}
& Bio2Token~\citep{liu2025biotoken}
& 0.644 & 0.983
& 0.693 & 0.981 \\

& BiHiTo~\citep{zheng2026bihito}
& 0.515 & 0.987
& 0.585 & 0.986 \\

\midrule

\multirow{6}{*}{\textbf{Residue-level}}
& DPLM-2~\citep{wang2025dplm}
& -- & 0.8860
& -- & 0.8644 \\

& \quad + cg2all (no fix)
& 3.4120 & 0.8859
& 4.5037 & 0.8644 \\

& \quad + PDBFixer
& 3.9657 & 0.8859
& 5.0213 & 0.8644 \\

& \quad + cg2all (fixed)
& 3.4358 & 0.8859
& 4.5288 & 0.8644 \\

& \quad + Rosetta fixbb
& 3.4918 & 0.8859
& 4.5931 & 0.8644 \\

& \textbf{ADT (ours)}
& 1.3157 & 0.9695
& 1.8261 & 0.9470 \\

\bottomrule
\end{tabular}%
}
\end{table}

\subsection{Comparison with Discrete Protein Tokenizers}
For completeness, we additionally compare three discrete protein tokenizers,
ESM3~\citep{hayes2025simulating}, DPLM-2~\citep{wang2025dplm}, and ADT,
on protein structure reconstruction. ESM3 is pretrained at substantially
larger scale, covering 2.78 billion protein sequences, whereas our tokenizer
is trained on only 330K structures from AFDB SwissProt. Nevertheless, ESM3
reports all-atom reconstruction RMSDs of approximately 1.30 and 1.80~\AA{}
on CASP14 and CASP15, respectively, compared with 1.32 and 1.83~\AA{} for
our method, indicating comparable all-atom reconstruction accuracy. Since
the publicly released ESM3 structure decoder exposes only backbone coordinates,
we restrict the directly evaluated comparison in
Table~\ref{app:casp_reconstruction} to backbone reconstruction.

More importantly, under direct evaluation, our method achieves lower backbone
RMSD and higher TM-score and GDT-TS than ESM3 on both CASP14 and CASP15,
while ESM3 achieves higher C$\alpha$-lDDT. This contrast indicates that ESM3
better preserves local backbone distances, whereas our method better preserves
global structural similarity. All metrics are averaged over individual proteins.

\begin{table}[t]
    \centering
    \caption{Protein structure reconstruction performance on CASP14 and CASP15.}
    \label{app:casp_reconstruction}
    \small
    \setlength{\tabcolsep}{3.5pt}
    \renewcommand{\arraystretch}{1.08}
    \resizebox{\linewidth}{!}{%
    \begin{tabular}{lccccc|ccccc}
        \toprule
        & \multicolumn{5}{c|}{\textbf{CASP14}}
        & \multicolumn{5}{c}{\textbf{CASP15}} \\
        \cmidrule(lr){2-6} \cmidrule(lr){7-11}
        \textbf{Method}
        & \textbf{C$\alpha$-lDDT} $\uparrow$
        & \textbf{BB RMSD} $\downarrow$
        & \textbf{TM-score} $\uparrow$
        & \textbf{GDT-TS} $\uparrow$
        & \textbf{GDT-HA} $\uparrow$
        & \textbf{C$\alpha$-lDDT} $\uparrow$
        & \textbf{BB RMSD} $\downarrow$
        & \textbf{TM-score} $\uparrow$
        & \textbf{GDT-TS} $\uparrow$
        & \textbf{GDT-HA} $\uparrow$ \\
        \midrule
        ESM3
        & 0.9653 & 1.9934 & 0.9552 & 0.9041 & 0.8068
        & 0.9672 & 2.0685 & 0.9455 & 0.8693 & 0.7470 \\

        DPLM-2
        & 0.8820 & 2.8576 & 0.8860 & 0.7718 & 0.5724
        & 0.8933 & 4.0223 & 0.8644 & 0.7032 & 0.5104 \\

        \textbf{ADT (Ours)}
        & 0.8970 & \textbf{0.9628} & \textbf{0.9695} & \textbf{0.9398} & 0.7686
        & 0.8972 & \textbf{1.5191} & \textbf{0.9470} & \textbf{0.9001} & 0.7285 \\
        \bottomrule
    \end{tabular}%
    }
\end{table}

\subsection{Unconditional Protein Generation}
To further evaluate the generative capability of ADT, we perform unconditional protein structure generation using a multimodal discrete diffusion model over amino-acid and ADT structural tokens, and compare with discrete protein generative models ESM3 and DPLM-2. Following prior work~\citep{ren2026sinae}, we generate 1,000 protein structures and evaluate designability (scRMSD $\leq 2.0$\AA), diversity, DPT, and novelty. As shown in Table~\ref{tab:protein_unconditional_generation}, ADT achieves the highest designability and diversity while maintaining competitive DPT and novelty. These results demonstrate that ADT supports diverse and designable protein generation across different discrete generative paradigms.

\begin{table}[t]
    \centering
    \caption{Unconditional protein structure generation.}
    \label{tab:protein_unconditional_generation}
    \small
    \setlength{\tabcolsep}{7pt}
    \renewcommand{\arraystretch}{1.1}
    \begin{tabular}{lcccc}
        \toprule
        \textbf{Method}
        & \textbf{Des.} $\uparrow$
        & \textbf{Div.} $\uparrow$
        & \textbf{DPT} $\downarrow$
        & \textbf{Nov.} $\uparrow$ \\
        \midrule
        ESM3
        & 0.61
        & 127
        & 0.37
        & \textbf{0.84} \\
        DPLM-2
        & 0.63
        & 130
        & 0.37
        & 0.72 \\
        ADT (Ours)
        & \textbf{0.68}
        & \textbf{139}
        & \textbf{0.35}
        & 0.76 \\
        \bottomrule
    \end{tabular}
\end{table}

\subsection{Qualitative Results}

\begin{figure}
    \centering
    \includegraphics[width=0.97\textwidth]{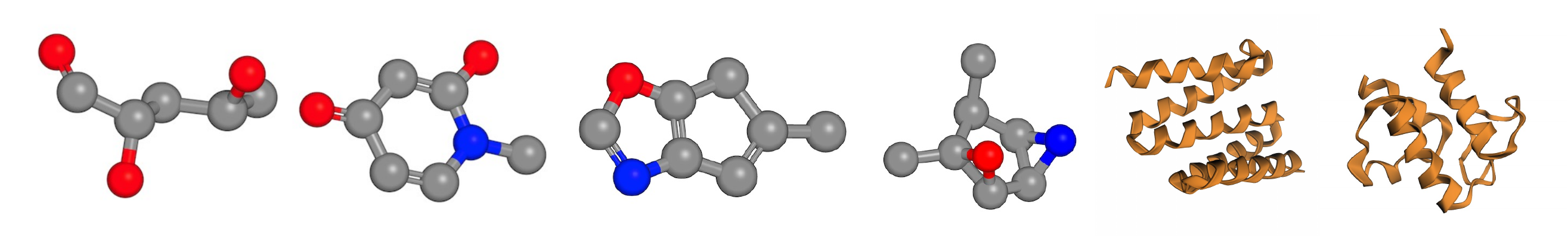}
    \caption{Qualitative results of samples generated by ADT for small molecules from QM9 and proteins.}
    \label{fig:qualitive}
\end{figure}

Figure~\ref{fig:qualitive} presents samples autoregressively generated
by ADT across different molecular domains. On QM9, ADT generates
diverse molecular topologies with multiple rings and complex bonding
patterns, including samples with 3 rings and 11 bonds. On
proteins, ADT scales to substantially larger structures and generates
distinct global folds with nontrivial three-dimensional conformations.
These results qualitatively demonstrate the capability of ADT to
autoregressively generate complex structured objects across different
structural scales and domains.

\section{Additional Ablation Studies}
\label{app:additional_ablations}

\subsection{Structural Information in Discrete Tokens}

To assess whether ADT tokens encode all-atom conformational information, we keep the amino-acid identities unchanged while randomly permuting the structural tokens within each protein, denoted as \emph{Shuffled tokens}. As shown in Table~\ref{app:token_shuffle}, shuffling the structural tokens substantially degrades reconstruction across all metrics, indicating that the decoder relies on residue-specific geometric information carried by the structural tokens.

\begin{table}[t]
\centering
\caption{Ablation of structural tokens on protein structure reconstruction.}
\label{app:token_shuffle}
\setlength{\tabcolsep}{3.5pt}
\renewcommand{\arraystretch}{1.1}
\small
\resizebox{\linewidth}{!}{%
\begin{tabular}{clcccccccc}
\toprule
& \textbf{Method}
& \textbf{C$\alpha$ lDDT} $\uparrow$
& \textbf{BB RMSD} $\downarrow$
& \textbf{AA lDDT} $\uparrow$
& \textbf{AA RMSD} $\downarrow$
& \textbf{SC RMSD} $\downarrow$
& \textbf{TM-score} $\uparrow$
& \textbf{GDT-TS} $\uparrow$
& \textbf{GDT-HA} $\uparrow$ \\
\midrule

\multirow{2}{*}{\textbf{CASP14}}
& Shuffled
& 0.3100
& 21.166
& 0.2978
& 21.709
& 22.278
& 0.2253
& 0.0373
& 0.0058 \\

& ADT
& \textbf{0.8970}
& \textbf{0.9628}
& \textbf{0.8223}
& \textbf{1.3157}
& \textbf{1.5938}
& \textbf{0.9695}
& \textbf{0.9398}
& \textbf{0.7686} \\

\midrule

\multirow{2}{*}{\textbf{CASP15}}
& Shuffled
& 0.3233
& 22.245
& 0.3103
& 22.733
& 23.274
& 0.2250
& 0.0354
& 0.0057 \\

& ADT
& \textbf{0.8972}
& \textbf{1.5191}
& \textbf{0.8258}
& \textbf{1.8261}
& \textbf{2.0779}
& \textbf{0.9470}
& \textbf{0.9001}
& \textbf{0.7285} \\

\bottomrule
\end{tabular}%
}
\end{table}

\subsection{Fine-grained Geometric Reconstruction on QM9}
We further evaluate fine-grained geometric fidelity in Table~\ref{tab:decoder_comparison}. Diffusion decoding reduces RMSD from 0.0584 to 0.0343 while also improving structure match, PoseBusters validity, and errors in bond lengths, bond angles, and dihedral angles. These results show that generative decoding more effectively reconstructs fine-grained geometry from the discrete representation than direct coordinate prediction.

\begin{table}[h]
\centering
\caption{Fine-grained Geometric Reconstruction on QM9.}
\label{tab:decoder_comparison}
\resizebox{\linewidth}{!}{%
\begin{tabular}{lcccccc}
\toprule
Decoder
& Match $\uparrow$
& RMSD  $\downarrow$
& PB Valid $\uparrow$
& Bond-length   $\downarrow$
& Bond-angle  $\downarrow$
& Dihedral  $\downarrow$ \\
\midrule
Transformer
& 98.9
& 0.0584
& 0.92
& 0.02219
& 1.4971
& 4.5234 \\
Diffusion
& \textbf{99.7}
& \textbf{0.0343}
& \textbf{0.93}
& \textbf{0.01557}
& \textbf{1.0929}
& \textbf{3.7622} \\
\bottomrule
\end{tabular}%
}
\end{table}

\subsection{Effect of Coarse-to-Fine Decoding}
To understand how residue-level representations can be progressively lifted to all-atom geometry, we investigate how decoder layers should be allocated between residue-level modeling and atom-level refinement.
Table~\ref{tab:coarse_to_fine} presents an ablation of our coarse-to-fine decoding design, where Coarse:Fine denotes the number of residue-level and atom-level decoder layers, respectively. The results yield two key observations:
1) Benefit of coarse-to-fine decoding. Coarse-to-fine configurations outperform both residue-only (12:0) and atom-only (0:12), showing that residue-level modeling and atom-level refinement provide complementary benefits for protein reconstruction.
2) Lightweight atomic refinement is sufficient. The 10:2 configuration consistently outperforms 8:4 and achieves the best performance across all reported metrics. This suggests that most structural modeling can be performed at the residue level, with only a lightweight atom-level stage needed for fine-grained refinement.

\begin{table}[h]
    \centering
    \caption{Ablation of coarse-to-fine diffusion decoding on protein structure reconstruction.}
    \label{tab:coarse_to_fine}
    \small
    \setlength{\tabcolsep}{3.5pt}
    \renewcommand{\arraystretch}{1.08}
    \resizebox{\linewidth}{!}{%
    \begin{tabular}{clccccccc}
        \toprule
        & \textbf{Coarse:Fine}
        & \textbf{C$\alpha$ lDDT} $\uparrow$
        & \textbf{BB RMSD} $\downarrow$
        & \textbf{AA lDDT} $\uparrow$
        & \textbf{AA RMSD} $\downarrow$
        & \textbf{TM-score} $\uparrow$
        & \textbf{GDT-TS} $\uparrow$
        & \textbf{GDT-HA} $\uparrow$ \\
        \midrule

        \multirow{4}{*}{\rotatebox[origin=c]{90}{\textbf{CASP14}}}
        & 12:0
        & 0.8568 & 1.1196 & 0.7934 & 1.5559
        & 0.9528 & 0.9027 & 0.7089 \\

        & 8:4
        & 0.8892 & 0.9876 & 0.8161 & 1.3574
        & 0.9668 & 0.9324 & 0.7571 \\

        & 10:2
        & \textbf{0.8970} & \textbf{0.9628}
        & \textbf{0.8223} & \textbf{1.3157}
        & \textbf{0.9695} & \textbf{0.9398}
        & \textbf{0.7686} \\

        & 0:12
        & 0.8624 & 1.0738 & 0.8067 & 1.5126
        & 0.9587 & 0.9068 & 0.7115 \\

        \midrule

        \multirow{4}{*}{\rotatebox[origin=c]{90}{\textbf{CASP15}}}
        & 12:0
        & 0.8574 & 1.7668 & 0.7967 & 2.1594
        & 0.9307 & 0.8630 & 0.6688 \\

        & 8:4
        & 0.8895 & 1.5583 & 0.8196 & 1.8840
        & 0.9444 & 0.8930 & 0.7170 \\

        & 10:2
        & \textbf{0.8972} & \textbf{1.5191}
        & \textbf{0.8258} & \textbf{1.8261}
        & \textbf{0.9470} & \textbf{0.9001}
        & \textbf{0.7285} \\

        & 0:12
        & 0.8629 & 1.6945 & 0.8101 & 2.0995
        & 0.9365 & 0.8671 & 0.6714 \\

        \bottomrule
    \end{tabular}%
    }
\end{table}

\subsection{Effect of Discrete Capacity}

To examine the effect of discrete capacity on residue-level structural
tokens, we vary the codebook size while keeping one structural token per
residue.
As shown in Table~\ref{tab:codebook_size}, reconstruction consistently
improves as the codebook grows from 2048 to 16384 across both CASP14 and
CASP15.
This trend suggests that increasing the representational capacity of each
structural token better preserves geometric information through the
discrete bottleneck.
A codebook of 8192 already provides strong reconstruction performance,
although increasing the vocabulary to 16384 yields further improvements.
We adopt 8192 as the default to balance reconstruction fidelity with
vocabulary size, given the larger prediction space that an expanded
vocabulary introduces for autoregressive modeling.

\begin{table}[t]
    \centering
    \caption{Effect of discrete codebook size on protein structure reconstruction.}
    \label{tab:codebook_size}
    \small
    \setlength{\tabcolsep}{3.5pt}
    \renewcommand{\arraystretch}{1.08}
    \resizebox{\linewidth}{!}{%
    \begin{tabular}{clccccccc}
        \toprule
        & \textbf{Codebook}
        & \textbf{C$\alpha$ lDDT} $\uparrow$
        & \textbf{BB RMSD} $\downarrow$
        & \textbf{AA lDDT} $\uparrow$
        & \textbf{AA RMSD} $\downarrow$
        & \textbf{TM-score} $\uparrow$
        & \textbf{GDT-TS} $\uparrow$
        & \textbf{GDT-HA} $\uparrow$ \\
        \midrule

        \multirow{4}{*}{\rotatebox[origin=c]{90}{\textbf{CASP14}}}
        & 2048
        & 0.8634 & 1.0342 & 0.7864 & 1.4734
        & 0.9563 & 0.9172 & 0.7243 \\

        & 4096
        & 0.8721 & 0.9818 & 0.8013 & 1.3869
        & 0.9634 & 0.9195 & 0.7431 \\

        & 8192
        & 0.8970 & 0.9628 & 0.8223 & 1.3157
        & 0.9695 & 0.9398 & 0.7686 \\

        & 16384
        & \textbf{0.9056} & \textbf{0.9467}
        & \textbf{0.8315} & \textbf{1.2768}
        & \textbf{0.9711} & \textbf{0.9584}
        & \textbf{0.7869} \\

        \midrule

        \multirow{4}{*}{\rotatebox[origin=c]{90}{\textbf{CASP15}}}
        & 2048
        & 0.8628 & 1.6324 & 0.7899 & 2.0413
        & 0.9338 & 0.8812 & 0.6853 \\

        & 4096
        & 0.8706 & 1.5587 & 0.8049 & 1.9284
        & 0.9409 & 0.8876 & 0.7041 \\

        & 8192
        & 0.8972 & 1.5191 & 0.8258 & 1.8261
        & 0.9470 & 0.9001 & 0.7285 \\

        & 16384
        & \textbf{0.9039} & \textbf{1.4636}
        & \textbf{0.8388} & \textbf{1.7382}
        & \textbf{0.9491} & \textbf{0.9173}
        & \textbf{0.7467} \\

        \bottomrule
    \end{tabular}%
    }
\end{table}

\subsection{Ablation of Autoregressive Factorization}

\begin{table}[h]
\centering
\caption{Effect of autoregressive factorization on QM9 generation.}
\label{tab:ar_generation}
\begin{tabular}{lcccc}
\toprule
Strategy & Valid $\uparrow$ & Unique $\uparrow$ & Div. $\uparrow$ & PB Valid $\uparrow$ \\
\midrule
Interleaved AR & 0.94 & 0.93 & 0.90 & 0.88 \\
G2C-AR         & 0.92 & 0.89 & 0.91 & 0.85 \\
C2G-AR         & \textbf{1.00} & \textbf{0.96} & \textbf{0.91} & \textbf{0.94} \\
\bottomrule
\end{tabular}
\end{table}

We examine the effect of autoregressive factorization on QM9 by comparing three generation strategies: interleaved autoregression, geometry-to-chemical autoregression (G2C-AR), and our Chemistry-to-Geometry autoregression (C2G-AR). All variants differ only in autoregressive factorization, with all other settings kept fixed. The interleaved AR follows $(s_1,c_1,\ldots,s_L,c_L)$; G2C-AR generates the complete structural sequence before the chemical sequence; and C2G-AR generates the complete chemical sequence before the structural sequence.

Table~\ref{tab:ar_generation} highlights two findings. First, C2G-AR consistently performs best, improving all metrics over interleaved AR and directly validating our Chemistry-to-Geometry factorization. Second, the gain is not a generic benefit of two-stage generation: reversing the order with G2C-AR reduces validity to 0.92 and PB validity to 0.85, below the interleaved baseline. These results isolate the benefit of generating structural tokens after the complete chemical sequence and directly support the C2G factorization used in our model.

\section{More Implementation Details}
\label{app:implementation}

\subsection{Tokenizer Model Architecture}
\label{app:model}
Both the small-molecule and protein tokenizers use pre-norm Transformer encoders, binary FSQ, and Diffusion Transformer (DiT) decoders, with 12 layers in both the encoder and decoder. The small-molecule tokenizer uses $d_{\mathrm{model}}=1536$, while the protein tokenizer uses $d_{\mathrm{model}}=256$. For proteins, the 12-layer DiT decoder consists of 10 residue-level layers followed by 2 intra-residue atom-level layers. Both tokenizers use 13-bit binary FSQ, corresponding to a codebook size of $8192$.

\subsection{Training Details}
\label{app:training}
Both the small-molecule and protein tokenizers are optimized using AdamW
with a learning rate of $3\times10^{-4}$. The mutual-information regularizer
is applied to both tokenizers with $\lambda_{\mathrm{MI}}=10^{-3}$. The temperature of mutual information $\tau$ is set to 1.0.

\paragraph{Small-molecule tokenizer.}
The small-molecule tokenizer is trained with the diffusion reconstruction
objective and the mutual-information regularizer.

\paragraph{Protein tokenize.}
For protein tokenizer training, we construct a high-confidence subset of AFDB Swiss-Prot by retaining monomeric proteins with 60–1,024 modeled residues and mean pLDDT above 85, resulting in 330,154 protein structures
In addition to the diffusion reconstruction objective and mutual-information
regularizer, the protein tokenizer uses auxiliary geometric losses.
Specifically, we define
\begin{equation*}
\begin{aligned}
\mathcal{L}_{\mathrm{bond}}
&=
\frac{1}{|\mathcal{B}|}
\sum_{(i,j)\in\mathcal{B}}
(\widehat d_{ij}-d_{ij})^2,
&
\mathcal{L}_{\mathrm{angle}}
&=
\frac{1}{|\mathcal{T}|}
\sum_{(i,j,k)\in\mathcal{T}}
(\cos\widehat\theta_{ijk}-\cos\theta_{ijk})^2,
\\
\mathcal{L}_{\mathrm{clash}}
&=
\frac{1}{|\mathcal{N}|}
\sum_{(i,j)\in\mathcal{N}}
[\max(0,\delta_{ij}-\widehat d_{ij})]^2,
&
\mathcal{L}_{\mathrm{dist}}
&=
\frac{1}{|\mathcal{P}|}
\sum_{(i,j)\in\mathcal{P}}
(\widehat d_{ij}^{\,\mathrm{C}_\alpha}
-d_{ij}^{\,\mathrm{C}_\alpha})^2.
\end{aligned}
\end{equation*}
Here, $\mathcal{B}$, $\mathcal{T}$, $\mathcal{N}$, and $\mathcal{P}$ denote
bonded atom pairs, bond-angle triplets, non-bonded atom pairs, and
$\mathrm{C}_\alpha$ pairs, respectively.
The weights for the backbone, bond-length, bond-angle, clash and distance are set to $1.0$, $0.1$, $0.1$, $0.1$ and $0.01$, respectively.

\subsection{Autoregressive Generation Details}
We train an unconditional autoregressive model on QM9 to evaluate the
generative capability of the learned discrete structural representation.
Each molecule is represented by a sequence of atom identities
$C=(c_1,\ldots,c_L)$ and the corresponding structural tokens
$S=(s_1,\ldots,s_L)$ obtained from the pretrained tokenizer.
Each atom is associated with one structural token. The tokenizer uses
13-bit binary FSQ, corresponding to a structural vocabulary of
$2^{13}=8192$ discrete codes. We set $\beta=1.0.$

\paragraph{Small-Molecule Autoregressive Generation.}
We use a 12-layer Transformer with a hidden dimension of 768, 12 attention heads, and a dropout rate of 0.1, trained from scratch. The autoregressive model is trained on the QM9 training split using AdamW with a batch size of 64 and a learning rate of $10^{-4}$. 
Teacher-forced prediction performance is evaluated on the validation set containing 12,981 molecules.

\paragraph{Unconditional Protein Generation.}
For unconditional protein generation, we train a multimodal discrete diffusion model over amino-acid and ADT structural tokens, following the generation paradigm of prior discrete protein models to enable a controlled comparison. The model consists of 33 bidirectional Transformer layers with a hidden dimension of 1280 and 20 attention heads. During training, amino-acid and structural tokens are corrupted according to an absorbing discrete diffusion process, and the model learns to recover the clean tokens from their corrupted counterparts. At inference, generation starts from masked amino-acid and structural tokens and iteratively denoises both modalities to jointly generate a protein sequence and its corresponding structural-token sequence. The generated structural tokens are decoded into all-atom coordinates by the frozen ADT diffusion decoder. The model is trained with a learning rate of $1\times10^{-4}$.

\paragraph{Sequence-Conditioned All-atom Protein Autoregressive Generation.}
\label{app:sequence-details}
The training set is constructed from the first 80 shards of AFDB v6 with pLDDT $\geq 90$, comprising approximately 32.2M structures and 10.1B residues after preprocessing.
We remove training structures with sequence identity greater than 40\% to any CAMEO22 target, requiring at least 80\% alignment coverage of the evaluation target.
Each training example consists of an amino-acid sequence paired with structural tokens obtained from the pretrained ADT tokenizer. 
We evaluate sequence-conditioned structure generation on 183 CAMEO22 targets~\citep{robin2021continuous}.

We use a  Qwen3.5-0.8B model as the backbone with a hidden dimension of 1024. 
During training, the amino-acid sequence is provided as context and the cross-entropy loss is applied only to the target structural-token sequence.
We optimize the model using AdamW with $\beta_1=0.9$, $\beta_2=0.95$, $\epsilon=10^{-8}$, and weight decay $0.1$. We use a global batch size of 2048, a peak learning rate of $10^{-4}$, and cosine learning-rate decay to $10^{-5}$, with a 5\% warmup period. Gradients are clipped to a maximum norm of 1.0.
At inference, given an amino-acid sequence, the model autoregressively generates structural tokens, which are then decoded into all-atom coordinates by the frozen ADT diffusion decoder.
We report greedy decoding in Table~\ref{tab:protein_generation_postprocess} for direct comparison with DPLM-2.

\subsection{Evaluation Metrics}
\label{app:metirc}
\subsubsection{Reconstruction Metrics}

\paragraph{RMSD.}
For a structure with $N$ atoms, ground-truth coordinates
$\mathbf{X}\in\mathbb{R}^{N\times3}$ and reconstructed coordinates
$\widehat{\mathbf{X}}\in\mathbb{R}^{N\times3}$, we report the
Kabsch-aligned RMSD:
\begin{equation*}
\operatorname{RMSD}(\mathbf{X},\widehat{\mathbf{X}})
=
\min_{\mathbf{R}\in SO(3),\,\mathbf{t}\in\mathbb{R}^{3}}
\frac{1}{\sqrt{N}}
\left\|
\mathbf{X}
-
\left(\widehat{\mathbf{X}}\mathbf{R}^{\top}
+\mathbf{1}\mathbf{t}^{\top}\right)
\right\|_{F}.
\end{equation*}
For proteins, RMSD is computed over the corresponding atom subset,
including backbone (BB RMSD), all atoms (AA RMSD), or side-chain atoms
(SC RMSD).

\paragraph{Structure Match Rate.}
For molecular reconstruction on QM9, a reconstruction is considered successfully
matched if a valid chemical and atomic correspondence can be established with
the reference molecule using a matching tolerance of $0.01$~\AA.
Match rate is the fraction of successfully matched
molecules, while RMSD is computed after optimal alignment and averaged over
the matched subset.

\paragraph{lDDT.}
The local Distance Difference Test (lDDT) measures the preservation of local
interatomic distances between the reconstructed and reference structures
without requiring global alignment. We report C$\alpha$-lDDT over C$\alpha$ atoms, BB-lDDT over backbone atoms, and AA-lDDT over all atoms.

\paragraph{TM-score.}
TM-score measures the global structural similarity between the reconstructed
and reference protein structures, with higher values indicating greater
agreement in the overall fold.

\paragraph{GDT.}
The Global Distance Test measures the fraction of residues that can be
superimposed within predefined distance thresholds. We report GDT-TS and
GDT-HA, with GDT-HA using stricter distance thresholds to assess
higher-accuracy structural agreement.

\subsubsection{Generation Metrics}

\paragraph{Valid.}
Fraction of generated molecules that satisfy chemical validity, including
valid valence assignments and a consistent molecular graph, as determined
by RDKit sanitization.

\paragraph{Unique.}
Fraction of valid generated molecules that are chemically distinct, measured
by unique canonical SMILES.

\paragraph{Novelty.}
Fraction of valid generated molecules that are not present in the training
set, determined by canonical SMILES.

\paragraph{Diversity.}
Mean pairwise molecular dissimilarity among valid generated molecules,
measured using Tanimoto similarity of Morgan fingerprints:
\[
\operatorname{Diversity}
=
1-
\frac{2}{N(N-1)}
\sum_{i<j}
\operatorname{Tanimoto}(\mathbf{f}_i,\mathbf{f}_j),
\]
where $\mathbf{f}_i$ denotes the Morgan fingerprint (radius 2, 2048 bits)
of molecule $i$.

\paragraph{PB Valid.}
Fraction of generated molecules that pass the PoseBusters validity checks,
providing a more stringent assessment of physicochemical plausibility through
chemical consistency, intra-molecular clashes, bond lengths, and bond angles.

\paragraph{Self-Consistency.}
For each generated protein structure, we design eight sequences using
ProteinMPNN and predict their structures with ESMFold. We report
\textbf{scRMSD} as the minimum Kabsch-aligned C$\alpha$ RMSD and
\textbf{scTM} as the maximum TM-score between the generated structure and
the predicted structures of its designed sequences.

\paragraph{Designability.}
Fraction of generated protein structures with
$\mathrm{scRMSD}\leq2.0\,\text{\AA}$, indicating that at least one designed
sequence folds consistently with the generated structure.

\paragraph{Protein Diversity.}
Number of distinct structural clusters among designable generated structures,
obtained by clustering with pairwise TM-score at a cutoff of $0.5$. Higher
values indicate broader structural coverage.

\paragraph{Protein Novelty.}
Structural dissimilarity of designable generated structures to known protein
structures, defined as
\[
\operatorname{Novelty}
=
\frac{1}{|\mathcal{D}|}
\sum_{\widehat{\mathbf{X}}\in\mathcal{D}}
\left[
1-
\max_{\mathbf{P}\in\mathrm{PDB}}
\operatorname{TM}(\widehat{\mathbf{X}},\mathbf{P})
\right],
\]
where $\mathcal{D}$ denotes the set of designable generated structures.
Higher values indicate greater structural deviation from the closest known
PDB structures.

\paragraph{Designable Pairwise TM-score (DPT).}
Designable Pairwise TM-score (DPT) measures the mean pairwise TM-score among designable samples (scRMSD $\leq 2.0~\AA$), where a lower value indicates greater structural diversity among valid generations.

\section{Limitations}
\label{app:limitations}
The focus of this work is to introduce and validate ADT as a compact discrete representation for all-atom molecular structures. To this end, we study two representative molecular modalities, small molecules and proteins, spanning atom-level and residue-level tokenization. A natural extension would be to expand ADT to other biomolecular modalities, such as nucleic acids and molecular complexes, toward a shared discrete representation across diverse molecular systems. Such a representation could enable joint autoregressive modeling of different molecular modalities and potentially benefit from structural knowledge transferred across them. While our experiments cover reconstruction, unconditional small-molecule generation, and sequence-conditioned protein structure generation, another promising direction is to incorporate functional or property conditioning for structure-based molecular design. We leave these directions for future investigation.

\section{Usage Of LLMs}
\label{app:llm}

Large language models were employed as an auxiliary tool to support manuscript preparation, including grammar checking, sentence refinement, and clarification of technical descriptions. All
AI-suggested text was carefully reviewed and revised by the authors to ensure accuracy, clarity, and
scientific integrity.
\end{document}